\pdfoutput=1

\documentclass[11pt]{article}

\usepackage[final]{coling}

\usepackage{times}
\usepackage{latexsym}

\usepackage[T1]{fontenc}

\usepackage[utf8]{inputenc}

\usepackage{microtype}

\usepackage{inconsolata}

\usepackage{graphicx}

\usepackage{booktabs}
\usepackage{array}
\usepackage{inconsolata}
\usepackage{tabularx}
\usepackage{float}
\usepackage{caption}
\usepackage{graphicx}
\usepackage{multirow}
\usepackage{setspace}
\usepackage{booktabs}
\usepackage{hyperref}
\usepackage{csquotes}
\usepackage{tabularx}
\usepackage{paralist}
\usepackage{float}

\title{Assessing the Human Likeness of AI-Generated Counterspeech}

\author{
  \textbf{Xiaoying Song\textsuperscript{1}}
  \textbf{Sujana Mamidisetty\textsuperscript{2}}
  \textbf{Eduardo Blanco\textsuperscript{3}}
  \textbf{Lingzi Hong\textsuperscript{1}}
\\
  \textsuperscript{1}College of Information, University of North Texas
\\
  \textsuperscript{2}TAMS, University of North Texas
\\
  \textsuperscript{3}Department of Computer Science, University of Arizona
\\
  \small{
  \{xiaoyingsong, sujanamamidisetty\}@my.unt.edu} \\ \small{eduardoblanco@arizona.edu lingzi.hong@unt.edu  
  }
}

\begin{document}
\maketitle
\begin{abstract}
Counterspeech is a targeted response to counteract and challenge abusive or hateful content.
It effectively curbs the spread of hatred and fosters constructive online communication. 
Previous studies have proposed different strategies for automatically generated counterspeech.
Evaluations, however, focus on relevance, surface form, and other shallow linguistic characteristics.
This paper investigates the human likeness of AI-generated counterspeech,
a critical factor influencing effectiveness.
We implement and evaluate several LLM-based generation strategies,
and discover that AI-generated and human-written counterspeech can be easily distinguished by both simple classifiers and humans. 
Further, we reveal differences in linguistic characteristics, politeness, and specificity. 
The dataset used in this study is publicly available for further research \footnote{\url{https://github.com/oliveeeee25/counterspeech_eval_humanlike}}.

\end{abstract}

\noindent
\textcolor{red}{Trigger warning: Read with caution. Examples in this paper may present toxic languages.}

\section{Introduction}

Hate speech is a form of online abuse that continues to be a significant public concern ~\cite{chung2023understanding}. 
Combating hate is crucial for fostering constructive online communications.
Counterspeech, a direct response to challenge, dispute, and neutralize hateful comments~\cite{benesch2014countering}, is one of the effective solutions~\cite{buerger2021counterspeech}. 

\begin{table}
\centering
\footnotesize
\begin{tabularx}{\columnwidth}{p{.4in} X}
\toprule
Hate Speech & If some racist feminist c*** who hates my country and actively worked to harm it by voting in migrant hordes, were on fire. I wouldn’t p*** on her to put it out, much less defend her from said migrants that she wanted here in the first place. \\
\midrule
Counter-speech &  I cannot agree with your statement that it is acceptable to harm or ignore the well-being of any group of people, including those who hold different political views. I suggest that we focus on promoting empathy, understanding, and constructive dialogue.\\
\bottomrule
\end{tabularx}
\caption{A hate speech post from Reddit and counterspeech generated by prompt with Llama2.
  We compare AI-generated and human-written counterspeech and conclude they are fundamentally different.
}
\label{Table: intro-example}
\end{table}


Recent studies have explored methods for generating counterspeech~\cite{qian2019benchmark,chung2021towards}. 
Human-written counterspeech is often contextually appropriate and can positively affect bystanders and encourage civil online interactions ~\cite{zhao2024risk}.
However, human involvement is time-consuming and costly. 
Models have been developed to automatically generate counterspeech~\cite{zhu2021generate, tekiroglu2022using,hong2024outcome}. The provision has the potential to mimic human-written counterspeech and assist in combating hate~\cite{bail2024can}.

However, counterspeech generation models may struggle to accurately understand semantic nuances, leading to misunderstandings and even backfiring~\cite{luger2016like, hadi2019humanizing, toader2019effect}. Sometimes they do not appropriately match the user's emotional state or the conversational context ~\cite{han2022chatbot, chung2021towards}. Table~\ref{Table: intro-example} shows an example of counterspeech generated by prompt with Llama2. The counterspeech is perceived to be robotic, as it ignores the hatred and anger directed at the victim in the language (``racist feminist c*** ...'') and instead focuses on criticizing the imagined scenario (``not put it out if she were on fire''). The final suggestion is vague and does not specifically address the harmful stereotypes in the original statement. 
It may lack nuanced understanding and the depth of empathy for effective communication~\cite{hangartner2021empathy}.

Human likeness is an important factor in crafting counterspeech. Studies find that more human-like counterspeech tends to be more effective~\cite{pinochet2024experimental}. The tone and form of counterspeech significantly affect the effectiveness in influencing behaviors of 
users~\cite{buerger2021counterspeech}. Generated counterspeech that embodies human characteristics including empathy, humor, and respect can foster emotional attachments~\cite{glikson2020human}, potentially making counterspeech more acceptable and reliable~\cite{jiang2023make}. 
Human-like responses are more likely to express understanding and social connection~\cite{go2019humanizing}, contribute to fostering genuine interactions, and build trust~\cite{fenwick2022importance}.

On the other hand, there are ethical concerns if AI-generated and human-written counterspeech are indistinguishable~\cite{hua2024limitations}. 
AI models may be used for manipulative purposes such as
swaying opinions not aligned with specific agendas~\cite{karnouskos2020artificial}. These models may also generate biased responses that could unintentionally reinforce stereotypes or cause harm~\cite{lucy2021gender, kenthapadi2023generative}. 
Evaluating the human likeness of AI-generated counterspeech is important
because it brings awareness to these issues and could help mitigate them.

Previous studies have focused on evaluating the relevance, quality, toxicity, persuasiveness, and effectiveness of counterspeech~\cite{saha2022countergedi, hong2024outcome}.
Some studied linguistic aspects using metrics such as BLEU, METEOR, BERTScore, GRUENV scores, type-token and distinct-n~\cite{zhu2021generate, tekiroglu2022using}.
However, few have conducted thorough evaluations regarding human likeness. 

We propose to assess the human likeness of counterspeech based on the distinguishability between AI-generated and human-written replies to hate speech. 
To understand factors related to human likeness, we further analyze the differences
between AI-generated and human-written counterspeech using linguistic factors, politeness, and specificity.
Polite responses are essential for productive online communications~\cite{hermoyo2023optimizing}. Specificity refers to targeted responses to hate speech that fit the conversation context, which is usually well captured by humans and vital for effective counterspeech~\cite {zheng2023makes}. 

This study addresses the following questions:
\begin{compactitem}
\item Which LLMs are better at mimicking human-written counterspeech?
\item Are there linguistic differences between AI-generated and human-written counterspeech?
\item Do AI-generated and human-written counterspeech differ in politeness and specificity?
\end{compactitem}

We implement several state-of-the-art counterspeech generation strategies with LLMs, 
including prompting for one counterspeech
(\textit{Prompt}), 
prompt for multiple replies and select the best
(\textit{Prompt and Select}), 
fine-tune LLMs with existing counterspeech datasets
(\textit{Fine-tune}), 
and outcome-constrained LLMs.
(\textit{Constrained}). 
We also collect human-written counterspeech,
including both examples from actual user-generated content (i.e., genuine content from Reddit)
and crowd workers (i.e., counterspeech generated on demand).

With AI-generated and human-written counterspeech, we build authorship attribution models to identify which AI models generate more human-like counterspeech.
The higher the classification accuracy, the less human-like the AI-generated counterspeech is. 
We further deploy a human annotation task to validate and check whether humans can discern the differences between AI-generated and human-written counterspeech. 
Additionally, we analyze linguistic characteristics, politeness, and specificity of the counterspeech generated by various methods to understand the differences amongst counterspeech depending on the source (types of AI-generated and human-written).
 
Our study shows both classifiers and humans can easily distinguish AI-generated and human-written counterspeech.
AI-generated counterspeech shows
(a) significant differences in linguistic characteristics,
and
(b) is more polite and less specific.

\begin{table*}[ht!]
	\small
	\centering
        \scalebox{0.73}{
	\begin{tabular}{l ccc ccc ccc}
		\toprule
		\multicolumn{1}{c}{\multirow{2}{*}{Prior work}} & \multicolumn{2}{c}{Dataset} & \multicolumn{3}{c}{Model}  & \multicolumn{1}{c}{\multirow{2}{*}{Analysis}} \\ 
		\cmidrule(lr){2-4} \cmidrule(lr){5-6} 
		& Domain & Size & Context? & Generation Model & Training Methods  &  \\ \midrule
		
		\citet{gagiano2023prompt}  & \begin{tabular}[c]{@{}c@{}}Translation; \\ News summarization; \\ Web text\end{tabular} & \begin{tabular}[c]{@{}c@{}}Human: 9,000; \\ AI: 9,000\end{tabular} & No & T5, GPT-X & Not specified & Authorship classification \\ \addlinespace
  
		\citet{zhou2023synthetic} & \begin{tabular}[c]{@{}c@{}}News writing; \\ Social media\end{tabular} & \begin{tabular}[c]{@{}c@{}}Human: 12,408; \\ AI: 500\end{tabular} & No & GPT-3 & Few-shot learning & \begin{tabular}[c]{@{}c@{}}Linguistic features; \\ Authorship classification\end{tabular} \\ \addlinespace
  
		\citet{an2023use}  & Scientific writing & \begin{tabular}[c]{@{}c@{}}Human: 400; \\ AI: 400\end{tabular} & No & ChatGPT & Fine-tuning & \begin{tabular}[c]{@{}c@{}}Linguistic features; \\ Similarity comparison; \\ Authorship classification\end{tabular} \\ \addlinespace
  
		\citet{buz2024investigating} & Social media & \begin{tabular}[c]{@{}c@{}}Human: 411,189; \\ AI: 6,000\end{tabular} & No & \begin{tabular}[c]{@{}c@{}}GPT-2, \\ GPT-Neo, \\ GPT-3.5-turbo\end{tabular} & \begin{tabular}[c]{@{}c@{}}Zero-shot learning; \\ Fine-tuning\end{tabular} & \begin{tabular}[c]{@{}c@{}}Linguistic features; \\ Human preference; \\ Authorship classification\end{tabular} \\ \addlinespace
  
		\citet{prajapati2024detection}  & \begin{tabular}[c]{@{}c@{}}Scientific writing; \\ Social media; QA\end{tabular} & \begin{tabular}[c]{@{}c@{}}Human: 30,482;\\ AI: 18,308\end{tabular} & Yes & ChatGPT-3 & Zero-shot learning & \begin{tabular}[c]{@{}c@{}}Authorship classification;\\ Statistical imbalance;\\ Linguistic features;\\ Fact verification\end{tabular} \\ \addlinespace
  
		Ours & \begin{tabular}[c]{@{}c@{}}Hate speech/counterspeech; \\ Social media\end{tabular} & \begin{tabular}[c]{@{}c@{}}Human: 29,181;\\ AI: 54,136\end{tabular} & Yes & Llama2 & \begin{tabular}[c]{@{}c@{}}Prompt; Prompt and Select; \\ Fine-tune; Constrained\end{tabular} & \begin{tabular}[c]{@{}c@{}}Human likeness;\\ Linguistic features;\\Politeness; Specificity\end{tabular} \\ 
		\bottomrule
	\end{tabular}}
	\caption{Comparison of differentiating AI-generated and human-written texts: an overview of previous research and our contributions. We are the first to explore several AI-generation strategies and target human likeness along with politeness and specificity.}
	\label{Table: Related work}
\end{table*}

\section{Related Work}

\noindent
\textbf{Counterspeech Generation} 
Several studies employ crowd workers or experts to curate counterspeech~\cite{chung2019conan, qian2019benchmark}. These human-written datasets have been widely utilized for training counterspeech models~\cite{tekiroglu2022using, saha2022countergedi,gupta2023counterspeeches}.
Hybrid approaches that combine human annotations and generative models have been proposed,
resulting in counterspeech corpora such as CONAN~\cite{chung2019conan} and multiCONAN~\cite{fanton2021human}.
Advanced generative models have been developed for a more sophisticated generation of counterspeech, for example, the \emph{prompt and select} method
has been shown to generate more diverse and relevant counterspeech~\cite{zhu2021generate}.
Other approaches include constraining models for generating polite~\cite{saha2022countergedi}, intent-based~\cite{gupta2023counterspeeches}, or outcome-oriented counterspeech~\cite{hong2024outcome}.
This study implements several representative models to generate counterspeech
and
conducts an extensive evaluation focusing on human likeness. 

\noindent
\textbf{Counterspeech Evaluation}
Counterspeech evaluations have focused on
relevance using BLEU, ROUGE 
~\cite{chung2019conan, qian2019benchmark, 
};
diversity with metrics like repetition rate 
~\cite{zhu2021generate, tekiroglu2022using};
and
linguistic quality using GRUENV metrics~\cite{jiang2023raucg}. 
These metrics offer insights limited to the lexical and semantic levels. 
Recently, researchers have developed evaluation metrics specifically tailored for counterspeech, such as politeness and effectiveness~\cite{saha2022countergedi, hong2024outcome}. 
However, few studies have investigated the human likeness of AI-generated counterspeech. 
This study fills this void by evaluating several generation methods.

\noindent
\textbf{Comparing AI-Generated and Human-Written Texts} 
Many studies have explored methods to differentiate AI-generated from human-written texts across domains and tasks.
Examples include translation~\cite{el2023ensemble}, news summarization~\cite{gagiano2023prompt}, scientific writing~\cite{ma2023abstract}, and social media posts~\cite{buz2024investigating}. Recent studies utilize datasets ranging from a few hundred to over 40,000 samples.
They employ models such as T5, GPT variants, and ChatGPT with zero-shot learning and fine-tuning. Table~\ref{Table: Related work} presents a summary of these studies and the differences with the work presented here.

Most curated datasets consist of texts without additional context 
\cite{el2023ensemble,buz2024investigating, an2023use}.
Few include context and involve sequences or groups of related texts~\cite{ji2024detecting, prajapati2024detection}. 
To distinguish between AI-generated and human-written texts,
many researchers conduct linguistic analysis and authorship attribution, supplemented with human evaluations~\cite{zhou2023synthetic,ma2023abstract,buz2024investigating,ji2024detecting, prajapati2024detection}.
Our study is among the few to work in a dialogue setting,
where texts are dialogue turns countering a hate speech---AI-generated following four strategies or human-written from two sources.

\section{Counterspeech Data}
\label{s:cs_curation}

Our experiments are grounded on large collections of hate speech posts and their counterspeech replies.
The curation strategy includes both human-written counterspeech and AI-generated.

\subsection{Human-written  Counterspeech} \label{ss:human_counterspeech}
Human-written counterspeech consists of replies to hateful content written by humans.
We consider counterspeech written by both
genuine Reddit users (i.e., users that post out of their own will)
and crowd workers who are tasked with writing counterspeech.
Genuine human-written counterspeech is collected from Reddit.
First, we use keyword- and community-based sampling methods to retrieve hate speech posts from 42 subreddits (Appendix A) with a higher prevalence of hate via the Pushshift API \footnote{\url{https://pushshift.io/api-parameters/}}. 
This step results in 27,491 hate speech posts and their replies.
Second, we identify replies that are counterspeech with three BERT classifiers individually fine-tuned with three existing counterspeech corpora~\cite{qian2019benchmark,chung2021towards, yu2022hate}. 
We consider a reply to be counterspeech if the three classifiers indicate so,
finally resulting in 14,973 (hate speech, counterspeech) pairs from Reddit. 

Counterspeech written by crowd workers is collected from the Benchmark dataset~\cite{qian2019benchmark}.
This corpus includes hate speech posts paired with counterspeech replies written by crowd workers on demand.
We identify 14,208 valid (hate speech, counterspeech) pairs in Benchmark.

\subsection{AI-generated Counterspeech}
\label{ss:ai_counterspeech}
We implement state-of-the-art strategies to generate counterspeech replies to hateful posts.
The starting point is the 29,181 (hate speech, counterspeech) pairs from Section \ref{ss:human_counterspeech}.
Note that 
(a) some strategies require only hate speech posts while others also require the counterspeech reply
and
(b)
all hate speech posts were written by real Reddit users.

We develop counterspeech generation models based on the following methods:
\begin{compactdesc}
\item[Prompt] LLMs are prompted to generate counterspeech given a hateful post~\cite{fraser2023makes, hassan2023discgen, saha2024zero}.
\item[Prompt and Select] LLMs are prompted to generate multiple counterspeech replies and classifiers are employed to select the most diverse and relevant one~\cite{zhu2021generate}.
\item[Fine-tune] LLMs are trained with (hate speech, counterspeech) pairs to learn to generate human-written counterspeech~\cite{chung2021towards,fanton2021human}.
\item[Constrained] We adopt reinforcement learning with LLMs for outcome-constrained generation~\cite{hong2024outcome}.
\end{compactdesc}
The last two strategies use the pairs described in Section \ref{ss:human_counterspeech}
combined with three existing corpora: CONAN \cite{chung2021towards}, MultiCONAN \cite{fanton2021human}, and Benchmark \cite{qian2019benchmark}.
We present details (specific prompts, hyperparameters, etc.) in Appendix B.
Ultimately, we obtain 54,136 AI-generated counterspeech replies
(14,799,  19,436 14,215, and 5,686 respectively, as LLMs sometimes refuse to complete the task).


\begin{table}
\centering
\scalebox{0.62}{
\begin{tabular}{lcccc}
\toprule
                  & Prompt & Prompt and Select & Constrained & Fine-tune \\ \midrule
Agreement Rate    & 97\%   & 97\%   & 97\%        & 94\%      \\
Cohen $\kappa$  & 0.94   & 0.94   & 0.93        & 0.88      \\ \bottomrule
\end{tabular}}
\caption{Inter-annotator agreements differentiating human-written and AI-generated counterspeech.
  The task is straightforward for humans}
\label{Table: human likeness reliability}
\end{table}

\section{Evaluation Methods}
We conduct evaluations comparing human-written and AI-generated counterspeech accounting for four major categories:
human likeness, linguistic differences, politeness, and specificity.

\paragraph{Human Likeness}
If it is easy to differentiate between human-written and AI-generated counterspeech, we can conclude that the latter is not human-like.
We develop BERT classifiers with our curated dataset (Section \ref{s:cs_curation}).
The dataset is divided into an 80/20 split for training and testing.
Specifically, we create five binary classifiers (human-written or AI-generated) depending on what is included in the AI-generated counterspeech:
 all or only the counterspeech generated by one of the four strategies (see Appendix B for details).

In addition, we conduct a human validation for deeper insights.
We randomly select 100 samples from each AI strategy ($4\times100)$ and human-written counterspeech ($2\times100)$,
combine them, and ask human annotators to label whether the counterspeech is human-written or AI-generated.
Two research assistants complete the annotation process.
Table~\ref{Table: human likeness reliability} presents the agreement rate and Cohen's Kappa score.
Both indicate high reliability, meaning that the task is straightforward for humans.

\paragraph{Linguistic Differences}
We use SEANCE~\cite{crossley2017sentiment} to analyze the linguistic features of counterspeech and conduct statistical tests to reveal the differences between human-written and AI-generated counterspeech.  
For each type of AI-based method, we randomly sample an equal number of human-written counterspeech for comparison. 
We utilize the Wilcoxon rank-sum test to discern significant distinctions between the generation of AI and humans. Additionally, we apply the Bonferroni correction to identify the most significant linguistic features~\cite{weisstein2004bonferroni}.

\begin{table}
\centering
\scalebox{0.50}{
\begin{tabular}{@{}ccccccc@{}}
\toprule
& \multicolumn{4}{c}{AI-generated} & \multicolumn{2}{c}{Human-written} \\ \cmidrule(lr){2-5} \cmidrule(lr){6-7}
                  & Prompt   & Prompt and Select   & Constrained   & Fine-tune   & User & Crowd \\ \midrule
Weighted Cohen $\kappa$ & 0.80 & 0.94 & 0.96 & 0.93 & 0.80 & 0.88   \\ \bottomrule
\end{tabular}}
\caption{Inter-annotator agreements assessing politeness of AI-generated and human-written counterspeech.}
\label{Table: politeness reliability}
\end{table}

\paragraph{Politeness}
The level of politeness assesses the degree of respectfulness and courtesy.
It provides a complementary perspective to understand the differences between AI-generated and human-written counterspeech.
We build a politeness prediction model,
a BERT model fine-tuned with the dataset by~\citet{saha2022countergedi}.
The model achieves an F1 score of 0.91.  
We use this model to predict the politeness level of the counterspeech replies on a scale of 0 to 7. A higher score indicates more politeness.

Additionally, we conduct a human validation to gain insights.
We randomly select 100 samples from each AI strategy ($4\times100)$ and human-written counterspeech ($2\times100)$,
combine them, and ask the same human annotators to label them on a scale from 0 to 7 (See Appendix C for the guidelines).
Given the challenge of achieving exact agreement on a 7-point scale,
the weighted Cohen's Kappa is employed to calculate inter-annotator agreement (Table \ref{Table: politeness reliability}).
The agreements ($\kappa \geq 0.8$) again indicate high reliability.
We calculate the average of the human-annotated politeness scores and conduct the Kruskal-Wallis test to explore whether there are significant differences in the politeness of human-written and AI-generated counterspeech.

\begin{table}
\centering
\scalebox{0.50}{
\begin{tabular}{@{}ccccccc@{}}
\toprule
& \multicolumn{4}{c}{AI-generated} & \multicolumn{2}{c}{Human-written} \\ \cmidrule(lr){2-5} \cmidrule(lr){6-7}
                  & Prompt & Prompt and Select & Constrained & Fine-tune & User & Crowd \\ \midrule
Weighted Cohen $\kappa$ & 0.87   & 0.86   & 0.80        & 0.95      & 0.93 & 0.81   \\ \bottomrule
\end{tabular}}
\caption{Inter-annotator agreements assessing the specificity of AI-generated and human-written counterspeech.}
\label{Table: suitableness reliability}
\end{table}

\paragraph{Specificity}
We conduct a human assessment of the specificity of a counterspeech reply with respect to the corresponding hate speech post.
This metric measures how well the counterspeech
(a) aligns with contextual information
and
(b) targets the topic in the hate speech post~\cite{tekiroglu2022using, jones2024multi}.
We have designed a Likert scale (1--5) to evaluate specificity,
where 5 indicates the highest specificity.
In order to assess reliability, we randomly select 100 samples from each counterspeech source (700 in total)
for annotators to assess.
Inter-annotator agreement is again very high (Table~\ref{Table: suitableness reliability}, $\kappa \geq 0.80$).

\section{Results}

\subsection{Human Likeness}

\begin{table}
\centering
\scalebox{0.59}{
\begin{tabular}{lccccccccc}
\toprule
\multirow{2}{*}{Strategy} & \multicolumn{3}{c}{AI}                 & \multicolumn{3}{c}{Human}                   & \multicolumn{3}{c}{Weighted Average}            \\ \cmidrule(lr){2-4}  \cmidrule(lr){5-7} \cmidrule(lr){8-10}
                            & P             & R             & F1             & P             & R             & F1            & P             & R             & F1            \\ \midrule
Prompt         & 0.99          & 1.00              & 0.99          & 1.00              & 0.99          & 0.99          & 0.99          & 0.99          & 0.99          \\ 
Prompt and Select              & 0.98          & 1.00              & 0.99          & 1.00              & 0.98          & 0.99          & 0.99          & 0.99          & 0.99          \\ 
Fine-tune      & 0.80  & 0.95 & \textbf{0.87} & 0.94 & 0.76 & \textbf{0.84} & 0.87 & 0.86 & \textbf{0.86} \\
Constrained                    & 0.99            & 1.00              & 1.00              & 1.00              & 0.99              & 1.00              & 1.00              & 1.00              & 1.00              \\ 

All           & 0.95 & 1.00     & 0.97 & 1.00     & 0.95 & 0.97 & 0.98 & 0.97 & 0.97 \\ \bottomrule
\end{tabular}}
\caption{Results of a BERT classifier differentiating
(a)~AI-generated counterspeech with each and all strategies
and
(b)~human-written counterspeech.
The high F1 scores indicate that AI-generated counterspeech is not human-like.
} 
\label{Table: Model Performanc1}
\end{table}

\textbf{Classifier-Based Results} Table~\ref{Table: Model Performanc1} shows the results of the BERT-based classifier tuned to differentiate human-written and AI-generated counterspeech using the four strategies. We draw two main conclusions.
First, the classifier performs well, achieving F1 scores above 0.97 for all strategies except \textit{Fine-tune}.
Second, the classifier struggles to distinguish counterspeech generated by \textit{Fine-tune} and human-written counterspeech (weighted average F1: 0.86).
These results indicate that the \textit{Fine-tune} model more closely resembles human-written counterspeech compared to other generation models.

\begin{table}
\centering
\scalebox{0.6}{

\begin{tabular}{@{}lccccccccc@{}}
\toprule
\multirow{2}{*}{Corpus} & \multicolumn{3}{c}{AI}                 & \multicolumn{3}{c}{Human}                   & \multicolumn{3}{c}{Weighted Average}            \\ \cmidrule(lr){2-4}  \cmidrule(lr){5-7} \cmidrule(lr){8-10}
                            & P             & R             & F1             & P             & R             & F1            & P             & R             & F1            \\ \midrule
CONAN                       & 0.97 & 0.98 & 0.97          & 0.98 & 0.96 & 0.97          & 0.97    & 0.97   & 0.97            \\
Gab                         & 0.86 & 0.99 & 0.92          & 0.99 & 0.84 & 0.91          & 0.93    & 0.92   & 0.92            \\
Reddit                     & 0.81 & 0.95 & \textbf{0.88} & 0.94 & 0.78 & \textbf{0.85} & 0.88    & 0.87   & \textbf{0.87}   \\
MultiCONAN                      & 0.92 & 0.99 & 0.95          & 0.99 & 0.91 & 0.95          & 0.95    & 0.95   & 0.95            \\
Effectiveness               & 0.96 & 0.98 & 0.97          & 0.98 & 0.96 & 0.97          & 0.97    & 0.97   & 0.97            \\ \bottomrule
\end{tabular}}

\caption{Results of a BERT classifier differentiating
(a)~AI-generated counterspeech with the \emph{Fine-tune} strategy using different corpora
and
(b)~human-written counterspeech.
The corpora include
CONAN~\cite{chung2019conan},
MultiCONAN~\cite{fanton2021human},
Gab and Reddit from Benchmark~\cite{qian2019benchmark},
and effective-oriented counterspeech~\cite{hong2024outcome}.}
\label{Table: Model Performanc2}
\end{table}

We further develop five classification models to investigate the differences when the \textit{Fine-tune} strategy uses different datasets. Table~\ref{Table: Model Performanc2} shows the results.
The counterspeech generated by models fine-tuned with the Reddit benchmark dataset presents the greatest challenge. The weighted F1 score (0.87) is lower than others, however, it is still high. 
This leads to the observation that fine-tuned LLMs more closely capture the style of human-written counterspeech in the target platform---Reddit in this study.
However, they are still limited in generating counterspeech that has a high human likeness.
Models trained with expert-generated (i.e., CONAN, MultiCONAN), or outcome-oriented (i.e., effectiveness) counterspeech are the easiest to differentiate (F1: 0.97, 0.95, and 0.97, respectively. 

\begin{table}
\centering
\scalebox{0.58}{
\begin{tabular}{llllllllll}
\toprule
\multicolumn{1}{c}{\multirow{2}{*}{Strategies}} & \multicolumn{3}{c}{AI} & \multicolumn{3}{c}{Human} & \multicolumn{3}{c}{Weight Average} \\  \cmidrule(lr){2-4}  \cmidrule(lr){5-7} \cmidrule(lr){8-10}
\multicolumn{1}{c}{}                           & P      & R     & F1    & P       & R      & F1     & P          & R         & F1        \\ \midrule
Prompt                                             & 0.91   & 1.00  & 0.95  & 1.00    & 0.93   & 0.97   & 0.96       & 0.96      & 0.96      \\
Prompt and Select                                             & 1.00   & 0.92  & 0.96  & 0.93    & 1.00   & 0.96   & 0.96       & 0.96      & 0.96      \\
Constrained                                             & 0.90   & 0.98  & 0.94  & 0.98    & 0.91   & 0.94   & 0.94       & 0.94      & 0.94      \\
Fine-tune                                             & 0.49   & 0.64  & \textbf{0.55}  & 0.63    & 0.48   & \textbf{0.55}   & 0.57       & 0.55      & \textbf{0.55}      \\ \bottomrule
\end{tabular}}
\caption{Results obtained by humans differentiating
(a)~AI-generated counterspeech with each strategy
and
(b)~human-written counterspeech.
Humans are much less reliable than the BERT classifier (Table \ref{Table: Model Performanc1}) identifying counterspeech obtained with the \emph{Fine-tune} strategy,
but otherwise are proficient.}
\label{Tabel: Human performance}
\end{table}

\noindent\textbf{Human Assessment}
Table~\ref{Tabel: Human performance}
presents the results when humans differentiate AI-generated and human-written counterspeech.
Counterspeech generated by \textit{Prompt}, \textit{Prompt and Select}, and \textit{Constrained} can be easily identified by human annotators.
The counterspeech by \textit{Fine-tune}, however, is more challenging to distinguish, a consistency mirrored in the computing-based evaluation.

Human performance is lower than model performance across all AI-generation methods.
We conduct an error analysis and find the following. 
First, Some human-written counterspeech is template-based, for example, ``Use of such language or words is not acceptable'' or ``Using that language doesn't help you make your point.'' Though counterspeech examples are human-written, their formulaic nature resembles robotic outputs, making it challenging for human annotators to distinguish them from AI-generated.
Second, \textit{Fine-tune} models can better mimic human-like responses, generating responses that closely resemble authentic replies on Reddit. 
This kind of generation mimics human emotion and language style, complicating the task of differentiating it from genuine human-written counterspeech.

\begin{table*}[ht]
\centering
\small
\begin{tabular}{@{}lccccc@{}}
\toprule
\textbf{Category}            & \textbf{Prompt}                  & \textbf{Prompt and Select}                  & \textbf{Constrained}             & \textbf{Fine-tune}               & \textbf{All}                     \\ \midrule
\multicolumn{6}{l}{\textbf{Textual factors}}                                                                                                                                                                \\
~~~~1st person pronouns          & $\downarrow\downarrow$           & $\uparrow\uparrow\uparrow$       & $\uparrow\uparrow\uparrow$       & $\uparrow\uparrow\uparrow$       & $\uparrow\uparrow\uparrow$       \\
~~~~Action words                 & $\downarrow\downarrow\downarrow$ & $\downarrow\downarrow\downarrow$ & $\downarrow\downarrow\downarrow$ & $\uparrow\uparrow\uparrow$       & $\downarrow\downarrow\downarrow$ \\
~~~~Format words                 & $\downarrow\downarrow\downarrow$ & $\downarrow\downarrow\downarrow$ & $\uparrow\uparrow\uparrow$       & $\downarrow\downarrow\downarrow$ & $\downarrow\downarrow\downarrow$ \\
~~~~Certainty words              & $\uparrow\uparrow\uparrow$       & $\uparrow\uparrow\uparrow$       & $\uparrow\uparrow\uparrow$       & $\uparrow\uparrow\uparrow$       & $\uparrow\uparrow\uparrow$       \\
~~~~Frequency words              & $\downarrow\downarrow\downarrow$ & $\downarrow\downarrow\downarrow$ & $\downarrow\downarrow\downarrow$ & $\downarrow\downarrow\downarrow$ & $\downarrow\downarrow\downarrow$ \\
~~~~Self-Expression words        & $\uparrow\uparrow\uparrow$       & $\uparrow\uparrow\uparrow$       & $\uparrow\uparrow\uparrow$       & $\downarrow\downarrow\downarrow$ & $\uparrow\uparrow\uparrow$       \\
~~~~Anticipation words           & $\uparrow\uparrow\uparrow$       & $\uparrow\uparrow\uparrow$       & $\uparrow\uparrow\uparrow$       & $\downarrow\downarrow\downarrow$ & $\uparrow\uparrow\uparrow$       \\
~~~~Overstated Words             & $\downarrow\downarrow\downarrow$ & $\downarrow\downarrow\downarrow$ & $\downarrow\downarrow\downarrow$ & $\uparrow\uparrow\uparrow$       & $\downarrow\downarrow\downarrow$ \\ \midrule
\multicolumn{6}{l}{\textbf{Emotional factors}}                                                                                                                                                              \\
~~~~Support and affiliation      & $\downarrow\downarrow\downarrow$ & $\downarrow\downarrow\downarrow$ & $\uparrow\uparrow\uparrow$       & $\downarrow\downarrow\downarrow$ & $\downarrow\downarrow\downarrow$ \\
~~~~Excite from pleasure or pain & $\downarrow\downarrow\downarrow$ & $\downarrow\downarrow\downarrow$ & $\downarrow\downarrow\downarrow$ & $\downarrow\downarrow\downarrow$ & $\downarrow\downarrow\downarrow$ \\
~~~~Negative                     & $\downarrow\downarrow\downarrow$ & $\downarrow\downarrow\downarrow$ & $\downarrow\downarrow\downarrow$ & $\downarrow\downarrow\downarrow$ & $\downarrow\downarrow\downarrow$ \\
~~~~Positive Words               & $\uparrow\uparrow\uparrow$       & $\uparrow\uparrow\uparrow$       & $\uparrow\uparrow\uparrow$       & $\downarrow\downarrow\downarrow$ & $\uparrow\uparrow\uparrow$       \\
~~~~Hatred                       & $\downarrow\downarrow\downarrow$ & $\downarrow\downarrow\downarrow$ & $\downarrow\downarrow\downarrow$ & $\downarrow\downarrow\downarrow$ & $\downarrow\downarrow\downarrow$ \\ \midrule
\multicolumn{6}{l}{\textbf{Social factors}}                                                                                                                                                                 \\
~~~~Religious words              & $\downarrow\downarrow\downarrow$ & $\downarrow\downarrow\downarrow$ & $\downarrow\downarrow\downarrow$ & $\downarrow\downarrow\downarrow$ & $\downarrow\downarrow\downarrow$ \\
~~~~Economic words               & $\uparrow\uparrow\uparrow$       & $\uparrow\uparrow\uparrow$       & $\uparrow\uparrow\uparrow$       & $\downarrow\downarrow\downarrow$ & $\uparrow\uparrow\uparrow$       \\
~~~~Respect                      & $\downarrow\downarrow\downarrow$ & $\downarrow\downarrow\downarrow$ & $\downarrow\downarrow\downarrow$ & $\downarrow\downarrow\downarrow$ & $\downarrow\downarrow\downarrow$ \\
~~~~Wealth                       & $\downarrow\downarrow\downarrow$ & $\downarrow\downarrow\downarrow$ & $\downarrow\downarrow\downarrow$ & $\downarrow\downarrow\downarrow$ & $\downarrow\downarrow\downarrow$ \\
~~~~Power                        & $\downarrow\downarrow\downarrow$ & $\downarrow\downarrow\downarrow$ & $\downarrow\downarrow\downarrow$ & $\downarrow\downarrow\downarrow$ & $\downarrow\downarrow\downarrow$ \\ \bottomrule
\end{tabular}
 \caption{Linguistic analysis comparing counterspeech generated by AI and humans across different AI-based generation methods. The up arrow indicates higher values in AI-generated counterspeech. The number of arrows indicates the p-value of the Wilcoxon rank-sum test (one: $p<0.05$, two: $p<0.01$, and three: $p<0.001$). All tests have passed Bonferroni correction.}
\label{Table: Linguistic analysis}
\end{table*}

\subsection{Linguistic Differences}
We summarize the findings into three factors: textual, emotional, and social in Table \ref{Table: Linguistic analysis}. Various linguistic differences exist between human-written and AI-generated counterspeech. The trend is consistent across most groups except \textit{Fine-tune}. Counterspeech generated by \textit{Fine-tune} tends to be more human-like, exhibiting distinct linguistic patterns compared to other AI-generated groups.

Textual factors refer to the language style, structure, and function. Human-written counterspeech tends to contain more words of action, format (except \textit{Constrained}), frequency, and overstated (except \textit{Fine-tune}). AI-generated counterspeech includes more 1st person pronouns (except \textit{Prompt}), self-expression words, and anticipation words (except \textit{Fine-tune}), which is consistent with the insight of ~\citet{munoz2023contrasting}. The use of certainty words is significantly higher in AI-generated counterspeech ~\cite{zhou2023synthetic},  and this trend is consistent across all groups. 

Regarding emotional factors, human-written counterspeech expresses stronger feelings, such as negative, hatred, and excitement. AI-generated counterspeech significantly conveys more positive emotion (except \textit{Fine-tune}).

Social factors are associated with social dynamics, norms, values, and structures, referring to the social context of the content. Most social factors are prevalent in the human-written counterspeech, including religious, respect, wealth, and power words. AI-generated counterspeech tends to contain more economic words (except \textit{Fine-tune}).

\begin{figure}
  \centering
  \includegraphics[width=0.45\textwidth]{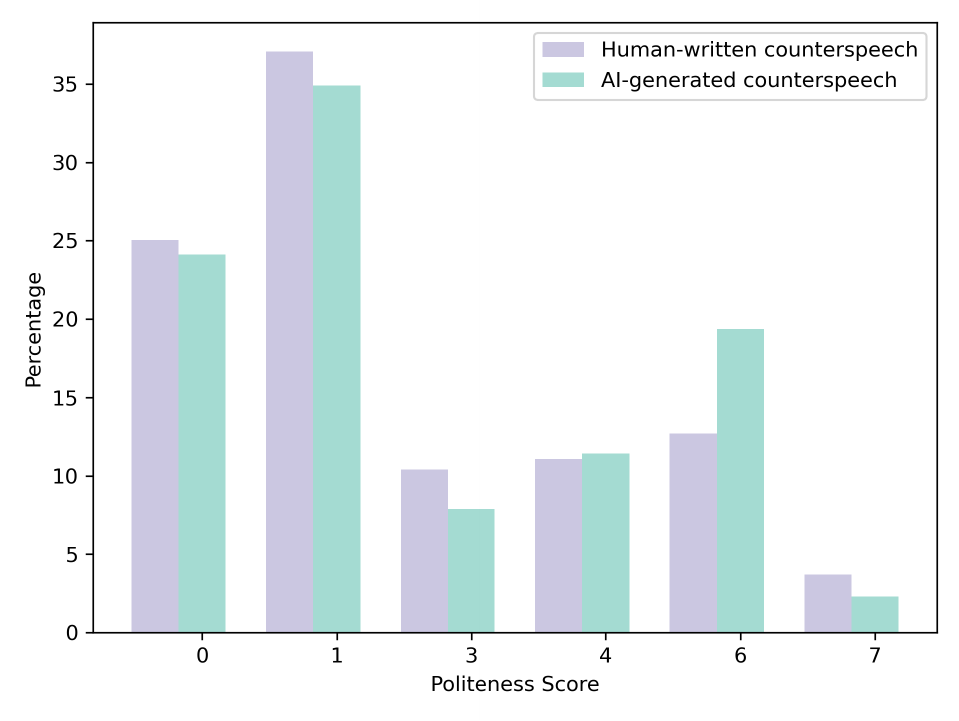}
  \caption{Politeness scores distribution of human-written and AI-generated counterspeech.
    Higher values indicate more politeness.
    AI-generated counterspeech is more polite.}
  \label{fig1}
\end{figure}

\subsection{Politeness}
\textbf{Classifier-Based Results} 
We conduct the Wilcoxon rank sum test between human-written and AI-generated counterspeech. The results show that the politeness of human-written counterspeech is significantly lower than AI-generated ($p<0.001$, Mean: 2.14 
 vs 2.37). 
Figure \ref{fig1} presents the politeness distribution of all the human-written and AI-generated counterspeech. 

Both types of generation have high density in lower politeness scores, with approximately 25\% at 0 and 35\% at 1. However, human-written counterspeech tends to concentrate on lower politeness levels than AI-generated counterspeech. AI-generated counterspeech has a higher proportion in higher politeness scores, especially at a score of 6 (human-written: 12\%, AI-generated: 20\%).


Figure~\ref{fig2}(a) shows the human annotation results of politeness. The results align with the findings of the evaluation using the classifier: AI-generated counterspeech demonstrates notably higher levels of politeness than human-written counterspeech. 
Responses by social media users are less polite than crowdsourced and AI-generated counterspeech.

In AI-generated methods, the \textit{Constrained} method generates more polite responses, with significantly higher polite scores according to the Kruskal-Wallis test ($p<0.001$). The counterspeech generated by \textit{Fine-tune} exhibits a notable spread towards both high and low ends, suggesting that counterspeech can range from very polite to impolite. 
Counterspeech in groups of \textit{Prompt} and \textit{Select} is less polite than that from \textit{Fine-tune} and \textit{Constrained}, but is still notably more polite than the user and crowdsourcing generation.

\subsection{Specificity}


\begin{figure*}
  \centering
    \includegraphics[width=1\textwidth]{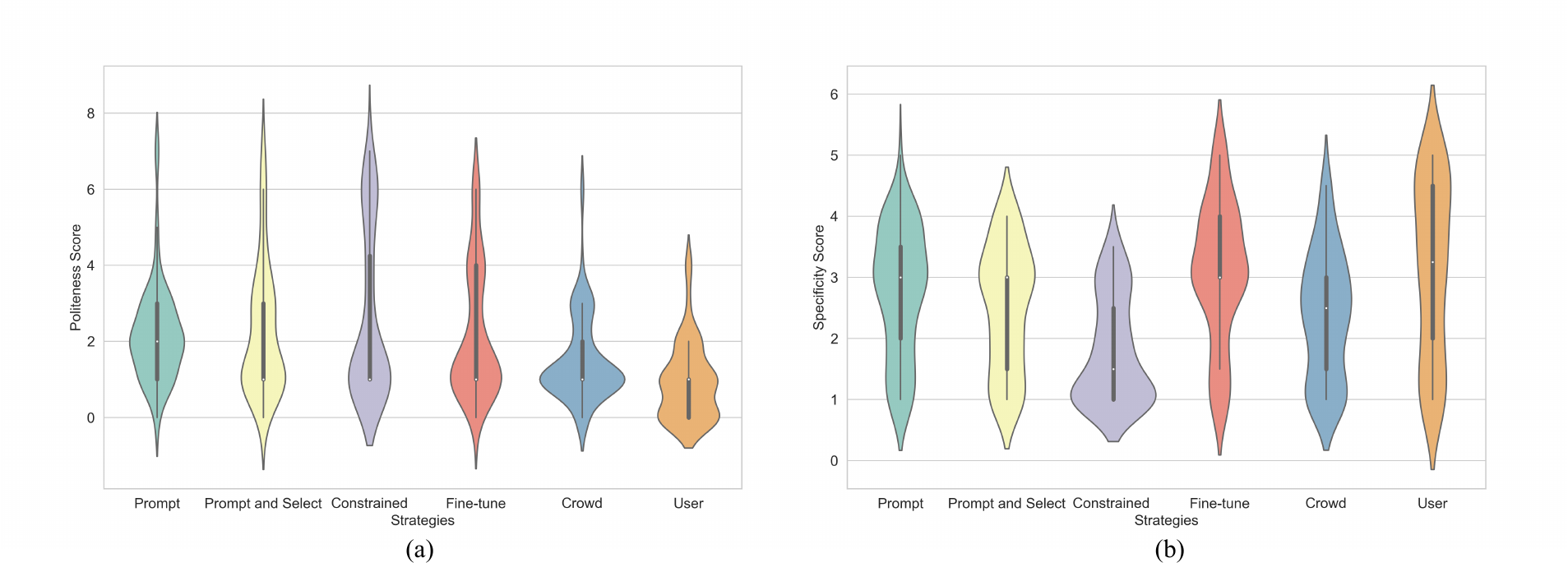}
  \caption{Distribution of politeness (left) and specificity (right) scores assigned by humans.
    We plot the distributions per strategy to generate counterspeech (four left-most plots) and human source (two right-most plots)).}
  \label{fig2}
\end{figure*}



The counterspeech generated by \textit{Prompt} and \textit{Prompt and Select} models exhibits a similar distribution with a moderate specificity level of 3 (Figure \ref{fig2}(b)). 
The \textit{Constrained} model-generated counterspeech has a median specificity score of 1.5, indicating that it lacks specificity and provides a more general reply. 
Comparatively, the counterspeech by \textit{Fine-tune} models shows high variances in the specificity scores, but most scores are between 3 and 4, indicating counterspeech is more targeted to the hate speech post at hand.

Crowdsourced counterspeech shows lower specificity(median score: 2.5). We find that a lot of crowdsourced counterspeech follows templates, uses vague wording, and does not directly address the specific hate speech post at hand. The user-generated counterspeech exhibits a broad distribution ranging from 1 to 5, with a higher frequency in scores of 4 and 5, suggesting some counterspeech has good specificity. Annotators also find that user-generated counterspeech is more contextually relevant and better targets the issues in a hateful post.

\section{Discussion}
According to evaluations in this study, AI-generated counterspeech can be easily differentiated by classification models and humans. Comparatively, \textit{Fine-tune} models outperform other automatic strategies in generating more human-like responses. These responses more closely resemble human-written counterspeech in linguistic features, exhibit a wide range of politeness levels, and have higher specificity scores. 

Counterspeech generated by \textit{Prompt} and \textit{Prompt and Select} tend to be long and lose efficacy. Some generic counterspeech (e.g., ``Finally, I would like to encourage you to engage in constructive conversation 
[\ldots]'') is commonplace. While the response is well-intentioned, it does not address the specific content of the hateful post.
Additionally, counterspeech by \textit{Prompt} and \textit{Prompt and Select} also shows template expressions such as ``It's understandable that [\ldots]'' and ``By working together to address these issues, we can build a society [\ldots]'', which are generic. 

Counterspeech by \textit{Constrained} is significantly more polite, with expressions like: `` I apologize [\ldots] I would encourage you to [\ldots]''. 
But there is a lot of repetitive content in the counterspeech, including ``I encourage you to strive for inclusivity, empathy, and respect for all people.'' Such responses seem template-based and can be easily distinguished from human-written counterspeech. 

In AI-generated counterspeech, we observe a common limitation: some counterspeech focuses on criticizing specific words and often includes misunderstandings. For example, terms such as ``retard'' are not always aimed at individuals with mental disabilities, yet counterspeech frequently responds with statements like ``Using a word that describes someone with a mental disability does not promote understanding.'' This type of counterspeech can be awkward or counterproductive. 

Crowdsourced counterspeech often follows a template, exhibiting a limited range of language styles. Many responses are not contextual and can be applied to various scenarios
(e.g., ``Use of such language or words is not acceptable.'')
It indicates that when humans are paid to generate counterspeech, they may repeat template-based replies that come across as repetitive or even robotic. 

User-written counterspeech demonstrates higher specificity, which can better capture the nuances in hate speech, offering more targeted and related responses, resonating the findings by ~\citet{go2019humanizing}. Since humans are better at interpreting human communication, they can more precisely grasp the intent behind hate speech~\cite{chen2018gunrock}. 
However, user-written counterspeech may be less polite. It is common for human response to convey natural emotions including anger, toxicity, and impoliteness when expressing opinions.

\section{Conclusion}
We propose to evaluate the human likeness of AI-generated counterspeech. 
We implement state-of-the-art counterspeech generation models following four strategies
(\textit{Prompt}, \textit{Prompt and Select}, \textit{Fine-tune}, and \textit{Constrained}), and use a Reddit hate speech / counterspeech dataset~\cite{qian2019benchmark} for counterspeech generation and evaluation. 
The human-written counterspeech comprises responses by crowd workers and social media users. 

We perform evaluations in authorship identification, linguistic features, politeness, and specificity. We find that counterspeech generated by current state-of-the-art models is distinguishable by both algorithms and humans. There are significant differences between AI-generated and human-written counterspeech in linguistic features, politeness, and specificity. AI-generated counterspeech is more polite and less focused, which are potential factors that make them differentiable from human-generated counterspeech. Fine-tuning LLMs with pertinent datasets makes the counterspeech generation more human-like. A future research direction might be the development of LLMs for generating more human-like replies. 

\section*{Limitations}
Our study has some limitations: (1) Not including all counterspeech generation methods. We focus on four popular generation methods (Prompt, Prompt and Select, Constrained, and Fine-tune) which are predominant in recent research. However, we do not explore other counterspeech generation strategies due to time and cost constraints. Nonetheless, our evaluation methods can be applied to assess other generation methods. 
(2) Limited evaluation scope. While our evaluation attempts to capture the nuances of human-written and AI-generated counterspeech, there are aspects such as tone and cultural sensitivity that are not evaluated, presenting avenues for future research.
(3) Subjectivity in human annotation. The process of annotating human likeness and politeness involves a degree of subjectivity that can lead to variability in results. We point out, however, that our human assessments follow standards for high reliability (double annotation and high agreements).
(4) Lack of cross-domain generalization. We focus on Reddit data, comparing AI-generated counterspeech with crowdsourcing counterspeech and user-generated counterspeech from Reddit. The results may vary depending on the nature of the original content and the platform on which it is applied. 
(5) Immediate research is needed.  As AI models are constantly improving, the performance of AI-generated results might change over time. In this study, we take the relatively new LLM Llama2 models for evaluation, however. More work needs to be done to identify whether our findings hold true for other models and newer versions.

\section*{Ethics Statement}
We fully consider the potential risks and benefits of our study. First, we collect data from Reddit, a public forum that makes data available to third parties. We have masked users' names and identities before analysis. Second, our study employs human annotators to evaluate counterspeech,  their names and identities are encrypted to avoid the identification of annotators. Third, we provide various AI-based methods to generate counterspeech that embody human likeness. There is a concern that some generated responses are indistinguishable from those created by humans, raising ethical considerations. We acknowledge the potential risks these methods may cause. However, the benefits will outweigh such risks, for example, using these generations in adversarial learning to develop more robust models for identifying AI-generated content.

\section*{Acknowledgements}
This work was supported by the Institute of Museum and Library Services (IMLS) National Leadership Grants under LG256661-OLS-24 and LG-256666-OLS-24.

\bibliography{custom}

\begin{thebibliography}{50}
\providecommand{\natexlab}[1]{#1}

\bibitem[{An et~al.(2023)An, Yang, Yang, and Wang}]{an2023use}
Ruopeng An, Yuyi Yang, Fan Yang, and Shanshan Wang. 2023.
\newblock Use prompt to differentiate text generated by chatgpt and humans.
\newblock \emph{Machine Learning with Applications}, 14:100497.

\bibitem[{Bail(2024)}]{bail2024can}
Christopher~A Bail. 2024.
\newblock Can generative ai improve social science?
\newblock \emph{Proceedings of the National Academy of Sciences}, 121(21):e2314021121.

\bibitem[{Benesch(2014)}]{benesch2014countering}
Susan Benesch. 2014.
\newblock Countering dangerous speech: New ideas for genocide prevention.
\newblock \emph{Available at SSRN 3686876}.

\bibitem[{Buerger(2021)}]{buerger2021counterspeech}
Catherine Buerger. 2021.
\newblock Counterspeech: A literature review.
\newblock \emph{Available at SSRN 4066882}.

\bibitem[{Buz et~al.(2024)Buz, Frost, Genchev, Schneider, Kaffee, and de~Melo}]{buz2024investigating}
Tolga Buz, Benjamin Frost, Nikola Genchev, Moritz Schneider, Lucie-Aim{\'e}e Kaffee, and Gerard de~Melo. 2024.
\newblock Investigating wit, creativity, and detectability of large language models in domain-specific writing style adaptation of reddit's showerthoughts.
\newblock \emph{arXiv preprint arXiv:2405.01660}.

\bibitem[{Chen et~al.(2018)Chen, Yu, Wen, Yang, Zhang, Zhou, Jesse, Chau, Bhowmick, Iyer et~al.}]{chen2018gunrock}
Chun-Yen Chen, Dian Yu, Weiming Wen, Yi~Mang Yang, Jiaping Zhang, Mingyang Zhou, Kevin Jesse, Austin Chau, Antara Bhowmick, Shreenath Iyer, et~al. 2018.
\newblock Gunrock: Building a human-like social bot by leveraging large scale real user data.
\newblock \emph{Alexa prize proceedings}.

\bibitem[{Chung et~al.(2023)Chung, Abercrombie, Enock, Bright, and Rieser}]{chung2023understanding}
Yi-Ling Chung, Gavin Abercrombie, Florence Enock, Jonathan Bright, and Verena Rieser. 2023.
\newblock Understanding counterspeech for online harm mitigation.
\newblock \emph{arXiv preprint arXiv:2307.04761}.

\bibitem[{Chung et~al.(2019)Chung, Kuzmenko, Tekiro{\u{g}}lu, and Guerini}]{chung2019conan}
Yi-Ling Chung, Elizaveta Kuzmenko, Serra~Sinem Tekiro{\u{g}}lu, and Marco Guerini. 2019.
\newblock Conan-counter narratives through nichesourcing: a multilingual dataset of responses to fight online hate speech.
\newblock In \emph{Proceedings of the 57th Annual Meeting of the Association for Computational Linguistics}, pages 2819--2829.

\bibitem[{Chung et~al.(2021)Chung, Tekiro{\u{g}}lu, Guerini et~al.}]{chung2021towards}
Yi-Ling Chung, Serra~Sinem Tekiro{\u{g}}lu, Marco Guerini, et~al. 2021.
\newblock Towards knowledge-grounded counter narrative generation for hate speech.
\newblock In \emph{Findings of the Association for Computational Linguistics: ACL-IJCNLP 2021}, pages 899--914. Association for Computational Linguistics.

\bibitem[{Crossley et~al.(2017)Crossley, Kyle, and McNamara}]{crossley2017sentiment}
Scott~A Crossley, Kristopher Kyle, and Danielle~S McNamara. 2017.
\newblock Sentiment analysis and social cognition engine (seance): An automatic tool for sentiment, social cognition, and social-order analysis.
\newblock \emph{Behavior research methods}, 49:803--821.

\bibitem[{El-Sayed and Nasr(2023)}]{el2023ensemble}
Ahmed El-Sayed and Omar Nasr. 2023.
\newblock An ensemble based approach to detecting llm-generated texts.
\newblock In \emph{Proceedings of the 21st Annual Workshop of the Australasian Language Technology Association}, pages 164--168.

\bibitem[{Fanton et~al.(2021)Fanton, Bonaldi, Tekiro{\u{g}}lu, and Guerini}]{fanton2021human}
Margherita Fanton, Helena Bonaldi, Serra~Sinem Tekiro{\u{g}}lu, and Marco Guerini. 2021.
\newblock Human-in-the-loop for data collection: a multi-target counter narrative dataset to fight online hate speech.
\newblock In \emph{Proceedings of the 59th Annual Meeting of the Association for Computational Linguistics and the 11th International Joint Conference on Natural Language Processing (Volume 1: Long Papers)}, pages 3226--3240.

\bibitem[{Fenwick and Molnar(2022)}]{fenwick2022importance}
Ali Fenwick and Gabor Molnar. 2022.
\newblock The importance of humanizing ai: using a behavioral lens to bridge the gaps between humans and machines.
\newblock \emph{Discover Artificial Intelligence}, 2(1):14.

\bibitem[{Fraser et~al.(2023)Fraser, Kiritchenko, Nejadgholi, and Kerkhof}]{fraser2023makes}
Kathleen~C Fraser, Svetlana Kiritchenko, Isar Nejadgholi, and Anna Kerkhof. 2023.
\newblock What makes a good counter-stereotype? evaluating strategies for automated responses to stereotypical text.
\newblock In \emph{Proceedings of the First Workshop on Social Influence in Conversations (SICon 2023)}, pages 25--38.

\bibitem[{Gagiano and Tian(2023)}]{gagiano2023prompt}
Rinaldo Gagiano and Lin Tian. 2023.
\newblock A prompt in the right direction: Prompt based classification of machine-generated text detection.
\newblock In \emph{Proceedings of the 21st Annual Workshop of the Australasian Language Technology Association}, pages 153--158.

\bibitem[{Glikson and Woolley(2020)}]{glikson2020human}
Ella Glikson and Anita~Williams Woolley. 2020.
\newblock Human trust in artificial intelligence: Review of empirical research.
\newblock \emph{Academy of Management Annals}, 14(2):627--660.

\bibitem[{Go and Sundar(2019{\natexlab{a}})}]{go2019humanizing}
Eun Go and S~Shyam Sundar. 2019{\natexlab{a}}.
\newblock Humanizing chatbots: The effects of visual, identity and conversational cues on humanness perceptions.
\newblock \emph{Computers in human behavior}, 97:304--316.

\bibitem[{Go and Sundar(2019{\natexlab{b}})}]{}
Eun Go and S~Shyam Sundar. 2019{\natexlab{b}}.
\newblock Humanizing chatbots: The effects of visual, identity and conversational cues on humanness perceptions.
\newblock \emph{Computers in human behavior}, 97:304--316.

\bibitem[{Gupta et~al.(2023)Gupta, Desai, Goel, Bandhakavi, Chakraborty, and Akhtar}]{gupta2023counterspeeches}
Rishabh Gupta, Shaily Desai, Manvi Goel, Anil Bandhakavi, Tanmoy Chakraborty, and Md~Shad Akhtar. 2023.
\newblock Counterspeeches up my sleeve! intent distribution learning and persistent fusion for intent-conditioned counterspeech generation.
\newblock \emph{arXiv preprint arXiv:2305.13776}.

\bibitem[{Hadi(2019)}]{hadi2019humanizing}
Rhonda Hadi. 2019.
\newblock When humanizing customer service chatbots might backfire.
\newblock \emph{NIM Marketing Intelligence Review}, 11(2):30--35.

\bibitem[{Han et~al.(2022)Han, Yin, and Zhang}]{han2022chatbot}
Elizabeth Han, Dezhi Yin, and Han Zhang. 2022.
\newblock Chatbot empathy in customer service: When it works and when it backfires.

\bibitem[{Hangartner et~al.(2021)Hangartner, Gennaro, Alasiri, Bahrich, Bornhoft, Boucher, Demirci, Derksen, Hall, Jochum et~al.}]{hangartner2021empathy}
Dominik Hangartner, Gloria Gennaro, Sary Alasiri, Nicholas Bahrich, Alexandra Bornhoft, Joseph Boucher, Buket~Buse Demirci, Laurenz Derksen, Aldo Hall, Matthias Jochum, et~al. 2021.
\newblock Empathy-based counterspeech can reduce racist hate speech in a social media field experiment.
\newblock \emph{Proceedings of the National Academy of Sciences}, 118(50):e2116310118.

\bibitem[{Hassan and Alikhani(2023)}]{hassan2023discgen}
Sabit Hassan and Malihe Alikhani. 2023.
\newblock Discgen: A framework for discourse-informed counterspeech generation.
\newblock In \emph{Proceedings of the 13th International Joint Conference on Natural Language Processing and the 3rd Conference of the Asia-Pacific Chapter of the Association for Computational Linguistics (Volume 1: Long Papers)}, pages 420--429.

\bibitem[{Hermoyo et~al.(2023)Hermoyo, Affandy et~al.}]{hermoyo2023optimizing}
R~Panji Hermoyo, Ali~Nuke Affandy, et~al. 2023.
\newblock Optimizing the use of polite language in responding to sexual harassment news on social media.
\newblock In \emph{1st UMSurabaya Multidisciplinary International Conference 2021 (MICon 2021)}, pages 392--401. Atlantis Press.

\bibitem[{Hong et~al.(2024)Hong, Luo, Blanco, and Song}]{hong2024outcome}
Lingzi Hong, Pengcheng Luo, Eduardo Blanco, and Xiaoying Song. 2024.
\newblock Outcome-constrained large language models for countering hate speech.
\newblock \emph{arXiv e-prints}, pages arXiv--2403.

\bibitem[{Hua et~al.(2024)Hua, Jin, and Jiang}]{hua2024limitations}
Shangying Hua, Shuangci Jin, and Shengyi Jiang. 2024.
\newblock The limitations and ethical considerations of chatgpt.
\newblock \emph{Data intelligence}, 6(1):201--239.

\bibitem[{Ji et~al.(2024)Ji, Li, Li, Guo, Qiu, Huang, Chen, Jiang, and Lu}]{ji2024detecting}
Jiazhou Ji, Ruizhe Li, Shujun Li, Jie Guo, Weidong Qiu, Zheng Huang, Chiyu Chen, Xiaoyu Jiang, and Xinru Lu. 2024.
\newblock Detecting machine-generated texts: Not just" ai vs humans" and explainability is complicated.
\newblock \emph{arXiv preprint arXiv:2406.18259}.

\bibitem[{Jiang et~al.(2023{\natexlab{a}})Jiang, Tang, Chen, Tanga, Wang, and Wang}]{jiang2023raucg}
Shuyu Jiang, Wenyi Tang, Xingshu Chen, Rui Tanga, Haizhou Wang, and Wenxian Wang. 2023{\natexlab{a}}.
\newblock Raucg: Retrieval-augmented unsupervised counter narrative generation for hate speech.
\newblock \emph{arXiv preprint arXiv:2310.05650}.

\bibitem[{Jiang et~al.(2023{\natexlab{b}})Jiang, Yang, and Zheng}]{jiang2023make}
Yi~Jiang, Xiangcheng Yang, and Tianqi Zheng. 2023{\natexlab{b}}.
\newblock Make chatbots more adaptive: Dual pathways linking human-like cues and tailored response to trust in interactions with chatbots.
\newblock \emph{Computers in Human Behavior}, 138:107485.

\bibitem[{Jones et~al.(2024)Jones, Mo, Fosler-Lussier, and Sun}]{jones2024multi}
Jaylen Jones, Lingbo Mo, Eric Fosler-Lussier, and Huan Sun. 2024.
\newblock A multi-aspect framework for counter narrative evaluation using large language models.
\newblock In \emph{Proceedings of the 2024 Conference of the North American Chapter of the Association for Computational Linguistics: Human Language Technologies (Volume 2: Short Papers)}, pages 147--168.

\bibitem[{Karnouskos(2020)}]{karnouskos2020artificial}
Stamatis Karnouskos. 2020.
\newblock Artificial intelligence in digital media: The era of deepfakes.
\newblock \emph{IEEE Transactions on Technology and Society}, 1(3):138--147.

\bibitem[{Kenthapadi et~al.(2023)Kenthapadi, Lakkaraju, and Rajani}]{kenthapadi2023generative}
Krishnaram Kenthapadi, Himabindu Lakkaraju, and Nazneen Rajani. 2023.
\newblock Generative ai meets responsible ai: Practical challenges and opportunities.
\newblock In \emph{Proceedings of the 29th ACM SIGKDD Conference on Knowledge Discovery and Data Mining}, pages 5805--5806.

\bibitem[{Lucy and Bamman(2021)}]{lucy2021gender}
Li~Lucy and David Bamman. 2021.
\newblock Gender and representation bias in gpt-3 generated stories.
\newblock In \emph{Proceedings of the third workshop on narrative understanding}, pages 48--55.

\bibitem[{Luger and Sellen(2016)}]{luger2016like}
Ewa Luger and Abigail Sellen. 2016.
\newblock " like having a really bad pa" the gulf between user expectation and experience of conversational agents.
\newblock In \emph{Proceedings of the 2016 CHI conference on human factors in computing systems}, pages 5286--5297.

\bibitem[{Ma et~al.(2023)Ma, Liu, Yi, Cheng, Huang, Lu, and Liu}]{ma2023abstract}
Yongqiang Ma, Jiawei Liu, Fan Yi, Qikai Cheng, Yong Huang, Wei Lu, and Xiaozhong Liu. 2023.
\newblock Is this abstract generated by ai? a research for the gap between ai-generated scientific text and human-written scientific text.
\newblock \emph{arXiv preprint arXiv:2301.10416}.

\bibitem[{Mu{\~n}oz-Ortiz et~al.(2023)Mu{\~n}oz-Ortiz, G{\'o}mez-Rodr{\'\i}guez, and Vilares}]{munoz2023contrasting}
Alberto Mu{\~n}oz-Ortiz, Carlos G{\'o}mez-Rodr{\'\i}guez, and David Vilares. 2023.
\newblock Contrasting linguistic patterns in human and llm-generated text.
\newblock \emph{arXiv preprint arXiv:2308.09067}.

\bibitem[{Pinochet et~al.(2024)Pinochet, de~Gois, Pardim, and Onusic}]{pinochet2024experimental}
Luis Hernan~Contreras Pinochet, Fernanda~Silva de~Gois, Vanessa~Itacaramby Pardim, and Luciana~Massaro Onusic. 2024.
\newblock Experimental study on the effect of adopting humanized and non-humanized chatbots on the factors measure the intensity of the user's perceived trust in the yellow september campaign.
\newblock \emph{Technological Forecasting and Social Change}, 204:123414.

\bibitem[{Prajapati et~al.(2024)Prajapati, Baliarsingh, Dora, Bhoi, Hota, and Mohanty}]{prajapati2024detection}
Manish Prajapati, Santos~Kumar Baliarsingh, Chinmayee Dora, Ashutosh Bhoi, Jhalak Hota, and Jasaswi~Prasad Mohanty. 2024.
\newblock Detection of ai-generated text using large language model.
\newblock In \emph{2024 International Conference on Emerging Systems and Intelligent Computing (ESIC)}, pages 735--740. IEEE.

\bibitem[{Qian et~al.(2019)Qian, Bethke, Liu, Belding, and Wang}]{qian2019benchmark}
Jing Qian, Anna Bethke, Yinyin Liu, Elizabeth Belding, and William~Yang Wang. 2019.
\newblock A benchmark dataset for learning to intervene in online hate speech.
\newblock In \emph{Proceedings of the 2019 Conference on Empirical Methods in Natural Language Processing and the 9th International Joint Conference on Natural Language Processing (EMNLP-IJCNLP)}, pages 4755--4764.

\bibitem[{Saha et~al.(2024)Saha, Agrawal, Jana, Biemann, and Mukherjee}]{saha2024zero}
Punyajoy Saha, Aalok Agrawal, Abhik Jana, Chris Biemann, and Animesh Mukherjee. 2024.
\newblock On zero-shot counterspeech generation by llms.
\newblock In \emph{Proceedings of the 2024 Joint International Conference on Computational Linguistics, Language Resources and Evaluation (LREC-COLING 2024)}, pages 12443--12454.

\bibitem[{Saha et~al.(2022)Saha, Singh, Kumar, Mathew, and Mukherjee}]{saha2022countergedi}
Punyajoy Saha, Kanishk Singh, Adarsh Kumar, Binny Mathew, and Animesh Mukherjee. 2022.
\newblock Countergedi: A controllable approach to generate polite, detoxified and emotional counterspeech.
\newblock \emph{arXiv preprint arXiv:2205.04304}.

\bibitem[{Tekiroglu et~al.(2022)Tekiroglu, Bonaldi, Fanton, Guerini et~al.}]{tekiroglu2022using}
Serra~Sinem Tekiroglu, Helena Bonaldi, Margherita Fanton, Marco Guerini, et~al. 2022.
\newblock Using pre-trained language models for producing counter narratives against hate speech: a comparative study.
\newblock In \emph{Findings of the Association for Computational Linguistics: ACL 2022}, pages 3099--3114. Association for Computational Linguistics.

\bibitem[{Toader et~al.(2019)Toader, Boca, Toader, M{\u{a}}celaru, Toader, Ighian, and R{\u{a}}dulescu}]{toader2019effect}
Diana-Cezara Toader, Grațiela Boca, Rita Toader, Mara M{\u{a}}celaru, Cezar Toader, Diana Ighian, and Adrian~T R{\u{a}}dulescu. 2019.
\newblock The effect of social presence and chatbot errors on trust.
\newblock \emph{Sustainability}, 12(1):256.

\bibitem[{Weisstein(2004)}]{weisstein2004bonferroni}
Eric~W Weisstein. 2004.
\newblock Bonferroni correction.
\newblock \emph{https://mathworld. wolfram. com/}.

\bibitem[{Yu et~al.(2022)Yu, Blanco, and Hong}]{yu2022hate}
Xinchen Yu, Eduardo Blanco, and Lingzi Hong. 2022.
\newblock Hate speech and counter speech detection: Conversational context does matter.
\newblock In \emph{Proceedings of the 2022 Conference of the North American Chapter of the Association for Computational Linguistics: Human Language Technologies}, pages 5918--5930.

\bibitem[{Yu et~al.(2024)Yu, Blanco, and Hong}]{yu2024hate}
Xinchen Yu, Eduardo Blanco, and Lingzi Hong. 2024.
\newblock Hate cannot drive out hate: Forecasting conversation incivility following replies to hate speech.
\newblock In \emph{Proceedings of the International AAAI Conference on Web and Social Media}, volume~18, pages 1740--1752.

\bibitem[{Zhao et~al.(2024)Zhao, Huang, Seligman, and Peng}]{zhao2024risk}
Yukun Zhao, Zhen Huang, Martin Seligman, and Kaiping Peng. 2024.
\newblock Risk and prosocial behavioural cues elicit human-like response patterns from ai chatbots.
\newblock \emph{Scientific reports}, 14(1):7095.

\bibitem[{Zheng et~al.(2023)Zheng, Ross, and Magdy}]{zheng2023makes}
Yi~Zheng, Bj{\"o}rn Ross, and Walid Magdy. 2023.
\newblock What makes good counterspeech? a comparison of generation approaches and evaluation metrics.
\newblock In \emph{Proceedings of the 1st Workshop on CounterSpeech for Online Abuse (CS4OA)}, pages 62--71.

\bibitem[{Zhou et~al.(2023)Zhou, Zhang, Luo, Parker, and De~Choudhury}]{zhou2023synthetic}
Jiawei Zhou, Yixuan Zhang, Qianni Luo, Andrea~G Parker, and Munmun De~Choudhury. 2023.
\newblock Synthetic lies: Understanding ai-generated misinformation and evaluating algorithmic and human solutions.
\newblock In \emph{Proceedings of the 2023 CHI Conference on Human Factors in Computing Systems}, pages 1--20.

\bibitem[{Zhu and Bhat(2021)}]{zhu2021generate}
Wanzheng Zhu and Suma Bhat. 2021.
\newblock Generate, prune, select: A pipeline for counterspeech generation$\backslash$$\backslash$against online hate speech.
\newblock \emph{Findings of the Association for Computational Linguistics}.

\end{thebibliography}

\appendix

\section{ Subreddit List}
\label{sec:appendix}
We collect Reddit data from the following 42 subreddits (Table \ref{Appendix Table 1}).

\begin{table}[H]
\centering
\scalebox{0.8}{
\begin{tabular}{@{}l@{}}
\toprule
 \multicolumn{1}{c}{Subreddits}                                                                                                                                                                                         \\ \midrule
 \begin{tabular}[c]{@{}l@{}} \textit{r/antiwork, r/changemyview, r/NoFap, r/Seduction,}\\ 

\textit{r/PurplePillDebate, r/ShitPoliticsSays, r/PurplePillDebate,}\\ 

\textit{r/bindingofisaac, r/FemaleDatingStrategy,}\textit{r/SubredditDrama}\end{tabular} \\

 \begin{tabular}[c]{@{}l@{}}\textit{r/KotakuInAction, r/DotA2, r/technology, r/modernwarfare,}\\ \textit{r/playrust, r/oblivion}\end{tabular}                                                                                             \\
 \begin{tabular}[c]{@{}l@{}}\textit{r/bakchodi, r/Feminism, r/PussyPass, r/MensRights,}\\ 

\textit{r/Sino, r/BlackPeopleTwitter, r/india, r/PussyPassDenied,} \\ \textit{r/TwoXChromosomes, r/GenZedong, r/antheism}\end{tabular}                   \\
 \begin{tabular}[c]{@{}l@{}}r/4Chan, \textit{r/justneckbeardthings, r/HermanCainAward,} \\ \textit{r/MetaCanada, r/DankMemes, r/ShitRedditSays}\end{tabular}                                                                              \\
 \begin{tabular}[c]{@{}l@{}} \textit{r/conspiracy,r/worldnews, r/Drama, r/TumblrInAction,}\\ \textit{r/lmGoingToHellForThis, r/TrueReddit.}\end{tabular}                                                                                  \\ \bottomrule
\end{tabular}}
\caption{Subreddit List.}
\label{Appendix Table 1}
\end{table}


\section{ Generation Details}
We implement several prominent generation methods in our study, categorized into four types: \textit{Prompt}, \textit{Select}, \textit{Constrained}, and \textit{Fine-tune}. We share the details of our experimental models.

\textbf{Model Parameters}
Due to limited computing resources, we use the Llama-2-7b-chat model for all generation experiments. This is to minimize the potential impact of using different LLMs on the generation results, thereby allowing the comparison to better reflect the differences in training methods. Specifically, the top k is set to be 8, the
temperature is 0.7, and the maximum length of reply is 512.

\textbf{Expertment details}
\textit{Prompt}: We request the Llama model to generate counterspeech based on our designed prompt in the following. This prompt is used in all generation methods.

\begin{displayquote}
    \texttt{System: "Generate a response in Reddit Style."}

    \texttt{User: "Here is the Reddit comment: <Hate Comment>. Please write a counterspeech to the Reddit hate comment."}
\end{displayquote}

The LLMs may avoid answering inappropriate questions and produce null results. We exclude null results in the evaluation. 

\textit{Select}: We request the model to generate 10 different responses for each HS, then implement the pruning and selection method by ~\citet{zhu2021generate} to generate counterspeech. 

\textit{Finetune}: We fine-tune the Llama-2-7b-chat model separately with the CONAN~\cite{chung2019conan}, MultiCONAN~\cite{fanton2021human}, Gab, and Reddit data from Benchmark~\cite{qian2019benchmark}. The \textit{Finetune} models can generate short responses. We remove responses with fewer than four words for further evaluation. 

\textit{Constrarined}: We incorporate reinforcement learning to contain the counterspeech generation following~\citet{hong2024outcome}. We design reward models to guide the reinforcement learning process: the constructive conversation outcome classifier~\cite{yu2024hate}. 
We first trained a finetuned model with the Reddit data from Benchmark to control the initial generation process, ensuring the model can generate responses to HS. Then the model is further trained with the reinforcement learning process, where responses that lead to low conversation incivility are awarded higher. The model after the reinforcement learning process is then used for counterspeech generation. 

\textbf{Computing Resources}
The computational resources applied in this research include a high-performance server equipped with an Intel Xeon Gold 6226R processor, 128 GB memory, and 3 Nvidia RTX 8000 GPUs.

\section{ Annoatation Details}
 We employ two PhD students with expertise in counterspeech study for the annotation tasks. Annotators were compensated on average with \$15 per hour. Annotators are females, one is a US citizen and one is an international student.

To maintain high-quality annotation in our study, we provide annotation training before human assessment. We have randomly collected 20 examples from the dataset for practice for each task. We require the annotators to review the HS first and then assess the corresponding counterspeech. After they annotate the data, we will check the labeled score and have a discussion with them to revise their understanding. After several rounds, their annotations show a higher consistency with expert annotation, indicating the great quality of the work. 

We also provide detailed annotation rubrics and examples to guide the human evaluation in Table~\ref{Appendix Table 2}. We have also designed the annotation rubric and examples for specificity in Table~\ref{Appendix Table 3}. Examples here may not represent the rubric very accurately, since it should be interpreted combined with HS. After two annotators complete their task, an expert reviews the results and makes final revisions. The final labels for human likeness are determined by the expert, while the final politeness and specificity scores are calculated as the average of the two annotators' labels.

\begin{table*}[t]
\centering
\large 
\scalebox{0.6}{
\begin{tabular}{@{}lll@{}}
\toprule
Score               & \multicolumn{1}{c}{Rubric}                                                                                                                       & \multicolumn{1}{c}{Example}                                                                                                                                                                                                          \\ \midrule
0-Very Impolite     & Counterspeech is perceived as rude or unfriendly                                                                                                 & The study was b**.                                                                                                                                                                                                                   \\ \addlinespace
1-Impolite          & Counterspeech that might not be overtly rude but lack warmth.                                                                                    & Use of the r-word is unacceptable                                                                                                                                                                                                    \\ 
\addlinespace
2-Somewhat Impolite & \begin{tabular}[c]{@{}l@{}}Counterspeech is neutral but could include language that might \\ be perceived as slightly unfriendly.\end{tabular}   & \begin{tabular}[c]{@{}l@{}}I don't think you're understanding what I'm saying.\end{tabular}                                                                                                                                        \\
\addlinespace
3-Neutral           & \begin{tabular}[c]{@{}l@{}}Counterspeech that is politely formulated without additional \\ warmth or friendliness but not impolite.\end{tabular} & \begin{tabular}[c]{@{}l@{}}He just doesn't have the mental capacity to understand \\the consequences of his actions\end{tabular}                                                                                                     \\
\addlinespace
4-Somewhat Polite   & \begin{tabular}[c]{@{}l@{}}Counterspeech that are polite and show an attempt at \\ engagement without significant warmth.\end{tabular}           & Please be careful with the words you choose!                                                                                                                                                                                         \\
\addlinespace
5-Polite            & \begin{tabular}[c]{@{}l@{}}Counterspeech that clearly make an effort to be \\ considerate and courteous.\end{tabular}                            & \begin{tabular}[c]{@{}l@{}}Instead, I suggest that you strive to promote inclusivity,\\ empathy, and understanding towards all individuals\end{tabular}                                                                          \\
\addlinespace
6-Very Polite       & \begin{tabular}[c]{@{}l@{}}Counterspeech includes expressions that enhance the \\ interaction positively.\end{tabular}                           & \begin{tabular}[c]{@{}l@{}}It's not okay to use derogatory language or slurs \\ to describe any group of people, Everyone deserves to \\ be treated with respect and dignity, regardless of their\\ beliefs or actions.\end{tabular} \\
\addlinespace
7-Extremely Polite  & Counterspeech is very thoughtful or considerate.                                                                                                 & \begin{tabular}[c]{@{}l@{}}Please refrain from using language that is \\ inappropriate or offensive, as it is not productive \\ or respectful. Thank you.  Is there something else \\I can help you with?\end{tabular}              \\ \bottomrule
\end{tabular}}
\caption{Politeness Rubric and Examples.}
\label{Appendix Table 2}
\end{table*}

\begin{table*}[ht]
\centering
\scalebox{0.5}{
\begin{tabular}{@{}lll@{}}
\toprule
Score                   & \multicolumn{1}{c}{Rubric}                                                                                                                        & \multicolumn{1}{c}{Example}                                                                                                                                                           \\ \midrule
1-Very Low Specificity  & \begin{tabular}[c]{@{}l@{}}Counterspeech is generic and does not address the specific content \\ or context of HS.\end{tabular}                   & It's offensive and hurtful to people with actual mental disabilities.                                                                                                                 \\
\addlinespace
2-Low Specificity       & Counterspeech addresses the general theme of HS but lacks details.                                                                                & Please refrain from using hateful ableist language in your posts                                                                                                                      \\
\addlinespace
3-Moderate Specificity  & \begin{tabular}[c]{@{}l@{}}Counterspeech somewhat addresses specific aspects of HS, \\ but may still include some general statements\end{tabular} & I think they mean that white people have an advantage in society.                                                                                                                     \\
\addlinespace
4-High Specificity      & \begin{tabular}[c]{@{}l@{}}Counterspeech directly engages with specific points HS, \\ providing targeted rebuttals.\end{tabular}                  & I'm not being retarded. I'm just not understanding why it's bad.                                                                                                                      \\
\addlinespace
5-Very High Specificity & Counterspeech is highly focused and precisely counters HS.                                                                                        & \begin{tabular}[c]{@{}l@{}}The government is not in the business of entitling people to free passes. \\ The government is in the business of protecting people's rights.\end{tabular} \\ \bottomrule
\end{tabular}}
\caption{Specificity Rubric and Examples.}
\label{Appendix Table 3}
\end{table*}

\end{document}